\documentclass[twoside,leqno,twocolumn]{article}

\usepackage[letterpaper]{geometry}

\usepackage{ltexpprt}
\usepackage{hyperref}

\usepackage{algorithm}
\usepackage[noend]{algpseudocode}
\usepackage{amsfonts}
\usepackage{amsmath}
\usepackage{array}
\usepackage{biblatex}
\usepackage{booktabs}
\usepackage[inline]{enumitem}
\usepackage[thinc]{esdiff}
\usepackage{graphicx}
\usepackage{multirow}
\usepackage{subcaption}
\usepackage[dvipsnames]{xcolor}
\usepackage{xspace}

\newcommand{\email}[1]{\href{mailto:#1}{\nolinkurl{#1}}}
\newcommand{\Kmeans}{$k$-means\xspace}
\newcommand{\Etal}{\textit{et al}\xspace}
\newcommand{\Eg}{\emph{e.g.},\xspace}
\newcommand{\Ie}{\emph{i.e.},\xspace}
\newcommand{\Etc}{\emph{etc.}\xspace}

\newcommand{\CD}{\mathit{CD}}
\newcommand{\ACD}{\mathit{ACD}}
\newcommand{\Green}[1]{\textcolor{ForestGreen}{#1}}
\newcommand{\Red}[1]{\textcolor{Red}{#1}}

\catcode`\%=12
\newcommand\pcnt{
\catcode`\%=14

\addbibresource{ms.bib}

\begin{document}

\newcommand\relatedversion{}

\title{\Large Cluster-level Group Representativity Fairness in \Kmeans Clustering}
\author{
    Stanley Simoes\thanks{School of EEECS, Queen's University Belfast, UK
    (\email{ssimoes01@qub.ac.uk})}
    \and
    Deepak P\thanks{School of EEECS, Queen's University Belfast, UK
    (\email{deepaksp@acm.org})}
    \and
    Muiris MacCarthaigh\thanks{School of HAPP, Queen's University Belfast, UK
    (\email{M.MacCarthaigh@qub.ac.uk})}
}

\date{}

\maketitle







\begin{abstract} \small\baselineskip=9pt
There has been much interest recently in developing fair clustering algorithms that seek to
do justice to the representation of groups defined along sensitive attributes such as \textit{race}
and \textit{gender}. We observe that clustering algorithms could generate clusters such that different
groups are disadvantaged within different clusters. We develop a clustering algorithm,
building upon the centroid clustering paradigm pioneered by classical algorithms such as
\Kmeans, where we focus on mitigating the unfairness experienced by the most-disadvantaged
group within each cluster. Our method uses an iterative optimisation paradigm whereby an
initial cluster assignment is modified by reassigning objects to clusters such that the
worst-off sensitive group within each cluster is benefitted. We demonstrate the effectiveness
of our method through extensive empirical evaluations over a novel evaluation metric on
real-world datasets. Specifically, we show that our method is effective in enhancing
cluster-level group representativity fairness significantly at low impact on cluster
coherence.
\end{abstract}

\subsection*{\small Keywords}
{\small algorithmic fairness, unsupervised machine learning, clustering, representativity}

\section{Introduction}\label{sec:introduction}

Fairness in ML~\cite{Chouldechova2020A} has seen much scholarly activity in recent times.
Within the broader umbrella of fair ML, several fair clustering methods have also been
developed~\cite{Chhabra2021An}. Most of these endeavours, starting from a pioneering work by
Chierichetti \Etal~\cite{Chierichetti2017Fair}, have considered ensuring proportional
representation of sensitive groups -- such as those defined on \textit{race} and \textit{gender} -- within each
cluster in the output; this is often referred to as {\it group
fairness}~\cite{Dwork2012Fairness}. Recent formulations of fair clustering are able to
incorporate considerations to fairness along multiple sensitive attributes
together~\cite{Abraham2020Fairness}. Such representational fairness of sensitive groups may be
seen as the application of the notion of proportional representation, {\it aka} statistical
parity~\cite{Besse2022A}, within clustering.

It may be argued that simply ensuring representation of sensitive groups within each cluster is
insufficient, especially within the popular paradigm of {\it centroid clustering}, pioneered by
the \Kmeans clustering algorithm~\cite{MacQueen1967Some}. In centroid clustering, each cluster
is represented by a prototype, often informally referred to as the centroid. A data object's
{\it proximity to its cluster centroid} is a key criterion determining the quality of the
clustering. For example, an object that happens to be very close to its centroid could be
thought of as being {\it `better represented'} within the clustering than another that happens
to be much further away from its centroid. Recent work on fair clustering has sought to deepen
uniformity of distance-to-centroid, dubbed as {\it representativity
fairness}~\cite{P2020Representativity}, across all dataset objects. Representativity fairness
seeks to ensure deeper levels of uniformity of representativity (\Ie centroid distance) across objects, regardless of
their sensitive attribute membership, and may thus be regarded as an instantiation of {\it
individual fairness}~\cite{Dwork2012Fairness}. The notion of representativity has been extended
to the framework of sensitive groups independently by Abbasi \Etal~\cite{Abbasi2021Fair}
and Ghadiri \Etal~\cite{Ghadiri2021Socially}. While the methodologies they propose are different, they are
strikingly similar in the nature of the fairness sought. They consider the aggregate
representativity (they use the term {\it representation}) across objects within each sensitive
group, and target proportionality along such group-specific aggregates across sensitive groups.
For example, within a social media profile clustering scenario, this implies that the mean
distance of female profiles from their centroids should be as close as possible to the mean
distance of male profiles from their centroids. This fairness condition targets that groups are
fairly treated across {\it all} the clusters in the output cluster assignment.

\begin{figure}[ht]
  \centering
    \includegraphics[width=0.7\linewidth]{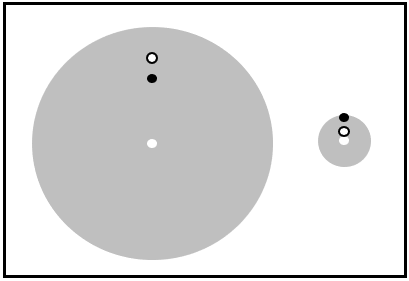}
    \caption{Illustration of representativity across differently sized clusters}
  \label{fig:rep-comparison}
\end{figure}

Against this backdrop, we note two issues with such representativity aggregations across
sensitive groups and clusters. {\it First}, note that {\it representativity} is quantified as
the distance of an object to {\it its} cluster centroid\footnote{Representativity, in a literal
sense, may be argued to mean centroid proximity than centroid distance; however, since it is
easier to deal with distances, we consistently refer minimisation of centroid distance as
representativity enhancement, which is essentially maximising centroid proximity.}. This
construction of representativity, we observe, does not yield well to cross-cluster comparisons,
especially when there could be clusters with widely varying sizes. Consider the toy 2-cluster
clustering in Figure~\ref{fig:rep-comparison} where the black and circled white objects
represent different sensitive groups, with the white object representing the centroid; also assume that the grey area is full of other objects
which we do not consider for now. The black object has better representativity (\Ie better centroid proximity) in the larger
cluster, and vice versa. However, given the size of the clusters, the same numeric difference in
representativity should be regarded as almost insignificant in the larger cluster, and very
consequential in the smaller cluster, when conditioned on the relative cluster sizes. In other
words, considering representativity against the backdrop of the cluster sizes, we may assume
that the circled white group gets a much better overall representativity than the black group.
In sharp contrast to such intuitive judgement, observe that the simple aggregate of
representativity misleadingly puts the black and circled white groups on an equal footing. {\it
Second}, note that the fairness goal seeks to bring about parity across sensitive groups. For a
clustering task, as in the case of earlier work on statistical parity for group fairness (\Eg
Chierichetti \Etal~\cite{Chierichetti2017Fair}, Abraham \Etal~\cite{Abraham2020Fairness}), the natural granularity for evaluating
clustering fairness is at the level of each cluster. The cross-cluster aggregation of
representativities is neither aligned with pragmatism (\Eg decision scenarios almost always
involve cluster-level decision-making~\cite{P2020Whither}) or underpinning political
justifications (\Eg in contrast to extant justifications of statistical
parity~\cite{Hertweck2021On}). We additionally note that the same sensitive group may be
advantaged in one cluster and disadvantaged in another. The paradigm of cross-cluster
aggregation of representativities allows such effects to cancel out, thus acting as a veneer to conceal potentially deep levels of
intra-cluster unfairness. 

\subsection{Our Contributions}

We propose a novel formulation of group fairness extending the notion of representativity
fairness along sensitive groupings in the data. Our {\it cluster-level group representativity
fairness} targets cluster-level fairness, thus mitigating identified cross-cluster
representativity aggregation issues. We propose a novel fair clustering method targeting to
optimise for cluster-level group representativity fairness, and illustrate, through extensive
empirical evaluations, that our method is able to achieve high degrees of fairness on
appropriate evaluation metrics.

\section{Related Work}

We now briefly summarise research on {\it group fairness} and {\it individual fairness} in fair
clustering. 

\subsection{Group Fairness} 

The notion of group fairness, which targets proportional representation of sensitive attribute
groups, was pioneered by Chierichetti \Etal~\cite{Chierichetti2017Fair} in 2017. Since then, group fair clustering research
has diversified into considering settings such as spectral
clustering~\cite{Kleindessner2019Guarantees}, hierarchical
clustering~\cite{Ahmadian2020FairHierarchical} and probabilistic scenarios~\cite{Esmaeili2020Probabilistic}.
Along another facet, variants of group fairness notions such as capped
representation~\cite{Ahmadian2019Clustering} and equitable group-level
representativity~\cite{Abbasi2021Fair, Ghadiri2021Socially} have been explored. There have also
been fairness conceptualisations such as {\it proportional
fairness}~\cite{Chen2019Proportionally} that straddle boundaries between group and individual
fairness. 

\subsection{Individual Fairness} 

At a risk of overgeneralisation, one may paraphrase individual fairness as ensuring that {\it
similar} objects be given similar outcomes (\Ie cluster membership in clustering). While this is
inarguably aligned with the clustering objective of maximising intra-cluster similarities and
minimising inter-cluster similarities, clustering algorithms can only achieve the objective on a
best-effort basis, making some shortfall inevitable. Specific formulations of individually fair
clustering methods have explored curtailing the shortfall through ways such as using
randomness~\cite{Brubach2020A}. P \& Abraham~\cite{P2020Representativity} considers optimising for the
uniformity of centroid-proximity ({\it aka} representativity) across objects. The focus on
representativity continues in much recent work, with Vakilian \& Yalciner~\cite{Vakilian2022Improved} considering an
object-specific upper bound for representativity and Chakrabarti \Etal~\cite{Chakrabarti2022A} proposing that
representativities between pairs of objects be bounded above by a specified multiplication factor.

\subsection{Positioning Our Work} 

Our notion of cluster-level group representativity fairness is a novel conceptualisation of
fairness unexplored in previous work. Our usage of cluster-level group fairness quantification as
an intermediate level between individual and group makes it distinct from previous work on both
fairness streams.

\section{Background}

We briefly outline the formulation of the popular \Kmeans clustering
algorithm~\cite{MacQueen1967Some}, as essential background to describe our method. Consider a set
of relational data objects $X = \{ \ldots x \ldots\}$ where $x \in \mathbb{R}^d$, which need to
be partitioned into $k$ clusters, $\mathcal{C} = \{ \ldots C \ldots\}$. \Kmeans uses an
EM-style~\cite{Dempster1977Maximum} optimisation framework to optimise for:
\begin{equation}\label{equation:kmeans}
    \sum_{C \in \mathcal{C}} \sum_{x \in C} ||x - \mu_C||^2
\end{equation}
where $\mu_C \in \mathbb{R}^d$ is the {\it centroid} or prototype object of cluster $C$. The
reader would notice that the objective relates to a given cluster assignment $\mathcal{C}$; the
EM-style optimisation starts with a given cluster assignment, and iteratively refines the cluster
assignment and centroids in order to minimise the objective in Equation~\ref{equation:kmeans}.

As a generic clustering method, \Kmeans is obviously agnostic to sensitive attribute groups to
which individual objects belong. Using the terminology of {\it representativity} (\Ie centroid
proximity), one may observe that \Kmeans seeks to optimise for the sum of representativities
across all objects in the dataset. This may be regarded as a
Benthamite~\cite{Bentham1996Collected} utilitarian objective that seeks {\it the greatest good
for the greatest number}. Abbasi \Etal~\cite{Abbasi2021Fair} and Ghadiri \Etal~\cite{Ghadiri2021Socially} note that this can result in cluster centroids representing
sensitive groups differently, often favouring one over the other.
This would have implications especially in cases where the centroids are used
to summarise objects in the clusters.
To mitigate the disparity in representativities of sensitive groups, Abbasi \Etal~\cite{Abbasi2021Fair} and
Ghadiri \Etal~\cite{Ghadiri2021Socially} independently proposed a new fair \Kmeans objective. Suppose each
data object in $X$ belongs to one of several groups $S$ (\Eg \textit{female}, \textit{male}, \Etc) defined across
a sensitive attribute $\mathcal{S} = \{ \ldots S \ldots\}$ (\Eg \textit{sex}). The fair \Kmeans
objective~\cite{Abbasi2021Fair, Ghadiri2021Socially} seeks to optimise for the worst-off
sensitive group at the cluster assignment level:
\begin{equation}\label{equation:fair-kmeans}
    \max_{S \in \mathcal{S}} \frac{1}{|S|} \sum_{C \in \mathcal{C}} \sum_{x \in C \cap S} ||x - \mu_C||^2
\end{equation}

The objective in Equation~\ref{equation:fair-kmeans} suffers from a variant of the issue observed in the
\Kmeans objective -- the averaging property can allow a sensitive group to have a low
representativity in one cluster which is discounted by moderately high representativities
in the other clusters. Further, a single sensitive group may not be the worst-off in all
clusters. Towards this, we turn our attention to mitigating group unfairness at the cluster
level, rather than the cluster assignment level.


\section{Problem Statement}\label{sec:problem-statement}

Our notion of cluster-level group representativity fairness targets to minimise the disparity in the
representativities of the sensitive groups within each cluster. We assume a single sensitive attribute
$\mathcal{S}$. Consider a cluster $C$ within a cluster assignment $\mathcal{C}$. The representativity loss
of the set of objects taking $S$ within $C$ is denoted as:
\begin{equation}\label{equation:group-repr}
    \overline{f}(S, C) = 
        \frac{1}{|C \cap S|} \sum\limits_{x \in C \cap S} ||x - \mu_C||^2
\end{equation}
Thus for each cluster $C$, we obtain a \emph{representativity vector} with one component per sensitive
group. We say that a cluster assignment is fair if all clusters in it are fair, and we say that a
cluster is fair if its corresponding representativity vector is uniform, \Ie the gap
between the best-off and worst-off groups is as small as possible.

In view of this novel formulation of cluster-level group representativity fairness, we consider
developing a metric towards quantifying adherence to the notion. In this paper, we look at the case
where the number of sensitive groups is not large. Accordingly, we capture the representativity
disparity between the best-off and worst-off groups simply by taking the difference in their representativities, given by:
\begin{equation}\label{equation:cluster-disparity}
    \CD(C) =
    \Bigg( \frac{\max\limits_{\substack{S \in \mathcal{S}}} \overline{f}(S,C)
          - \min\limits_{\substack{S \in \mathcal{S}}} \overline{f}(S,C)}
         {\min\limits_{\substack{S \in \mathcal{S}}} \overline{f}(S,C)} \Bigg) \times 100
\end{equation}
$\CD(C)$, or {\it cluster disparity}, quantifies the difference between the representativity losses for the best-off and worst-off groups within cluster $C$, expressed as a percentage. Evidently, we would ideally like $\CD(C)$ to be $0.0$ which is achieved when the same representativity loss is experienced by each sensitive group. The formulation of $\CD(C)$ as a percentage allows a fair basis for aggregating this across clusters of different sizes and shapes, thus addressing the first issue with representativity aggregation pointed out in Section~\ref{sec:introduction} -- that of cluster sizes. We aggregate this across clusters in the cluster assignment $\mathcal{C}$ to arrive at a single measure for the overall cluster assignment:
\begin{equation}\label{equation:acd}
    \ACD(\mathcal{C}) = \frac{1}{|\mathcal{C}|} \sum_{C \in \mathcal{C}} \CD(C)
\end{equation}
The {\it average cluster disparity} abbreviated as $\ACD(\mathcal{C})$, being an average of percentages, evaluates to a non-negative value, with a lower value indicating a smaller disparity
between the worst-off group and best-off group in individual
clusters, and consequently a fairer clustering.

\section{Proposed Method}

We now describe our fair clustering method focused on cluster-level group representativity fairness. We start by describing our objective function followed by the optimisation framework. 

\subsection{Objective Function}

The notion of fairness outlined in Section~\ref{sec:problem-statement} targets to enhance the
uniformity of the representativity vector for each cluster. This can be trivially achieved by having a
high representativity loss for all sensitive groups (as noted by Abbasi \Etal~\cite{Abbasi2021Fair}) which would
make the resulting cluster assignment of poor utility. Towards improving uniformity of the representativities, we take cue from contemporary theories in political philosophy and focus on mitigating
the representativity loss (\Ie  $\overline{f}(.,.)$) experienced by {\it the
worst-off group within each cluster}. The worst-off group in a cluster $C$ is given by:
\begin{equation}\label{equation:worstoff-group}
    \mathop{\arg\max}_{S \in \mathcal{S}} \ \ \overline{f}(S, C)
\end{equation}
This notion espouses the ethos across several popular
philosophical theories including the {\it concern for the most vulnerable} within the famed difference 
principle~\cite{Freeman2018Rawls} of distributive justice due to John
Rawls\footnote{\url{https://en.wikipedia.org/wiki/John\_Rawls}}.

Thus, the overall clustering objective we address here is to generate a cluster assignment that comprises coherent clusters (the singular focus of algorithms such as \Kmeans) where, additionally, the {\it representativity loss for the worst-off group within each cluster} is mitigated as much as possible. Given that the utilitarian consideration of {\it cluster coherence} (one that classical \Kmeans also targets to optimise) would be in apparent tension with the {\it cluster-level group representativity fairness} consideration, we look to deepen the latter at as little detriment to the former as possible. 

Given our intent of mitigating the representativity loss of the worst-off sensitive group within each cluster, we model our objective function as simply the aggregate of representativity losses of the worst-off cluster-level group:
\begin{equation}\label{equation:group-objective}
    J(\mathcal{C}) = \sum_{C \in \mathcal{C}} \max_{S \in \mathcal{S}} \bigg( \frac{1}{|C \cap S|} \sum\limits_{x \in C \cap S} ||x - \mu_C||^2 \bigg)
\end{equation}
This, as may be noted, captures the fairness ethos espoused by our formulation of cluster-level group representativity fairness, albeit using significantly different methodology. 


\subsection{Optimisation Framework}

To estimate the parameters -- $\mathcal{C}$ and $\{\ldots \mu_C \ldots \}$ -- in our objective function
(Equation~\ref{equation:group-objective}), we follow the same EM-style iterative
procedure as used in the classical \Kmeans algorithm that alternates
between the following E and M steps:
\begin{enumerate}[leftmargin=*]
    \item {\bf E-step:} estimate the cluster assignment for each $x \in X$ keeping the
    set of cluster centroids $\{\ldots \mu_C \ldots \}$ fixed, and
    \item {\bf M-step:} estimate the set of cluster centroids $\{\ldots \mu_C \ldots \}$ keeping the cluster assignment stationary.
\end{enumerate}
The method we propose falls in the category of post-processing methods which is in line with
classical works in fair clustering such as Ahmadian \Etal~\cite{Ahmadian2019Clustering}, Bera \Etal~\cite{Bera2019Fair}, Esmaeili \Etal~\cite{Esmaeili2020Probabilistic}, Kleindessner \Etal~\cite{Kleindessner2019Fair}, among others. It operates over the output of a utilitarian
clustering algorithm -- the classical \Kmeans algorithm; we use the generated cluster centroids as
the initial cluster centroids for our method. We now describe the details of the E and M steps
within our optimisation framework.

\subsection{E-step: Estimating the Cluster Assignment}

Given the set of cluster centroids $\{\ldots \mu_C \ldots \}$, we need to assign objects to clusters that minimise the objective function in Equation~\ref{equation:group-objective}. Towards operationalising this, given the complexity of the objective function, we perturb the existing cluster assignment by greedily reassigning objects -- in round-robin fashion -- to new clusters such that the value of the objective function decreases. Thus, if an object $x \in X$ is reassigned from cluster $C$ to cluster $C'$, the new cluster assignment $\mathcal{C}'$ is given by:
\begin{equation}\label{equation:cluster-update}
    \mathcal{C}' =
    \mathcal{C}
    \setminus \big\{C, C'\big\}
    \cup \big\{C \setminus \{x\}, C' \cup \{x\}\big\}
\end{equation}

Algorithm~\ref{algo:reassign} outlines our greedy approach. Within each E-step, this entails trying out $\mathcal{O}(|X| \times k)$ cluster reassignments, $k-1$ per object. It is notable that the change between $J(\mathcal{C}')$ and $J(\mathcal{C})$ where $\mathcal{C}$ and $\mathcal{C}'$ differ in the membership of a single object can be computed very efficiently without a full dataset-wide estimation. While we do not include the details of such efficient computations herein, the computation of such incremental changes is similar in spirit to what is outlined within Section 4.2.1 in Abraham \Etal~\cite{Abraham2020Fairness}.

\begin{algorithm}[ht]
    \caption{Reassign \label{algo:reassign}}
    \begin{algorithmic}[1]
        \ForAll {$x \in X$}
            \ForAll {$C' \in \mathcal{C}$}
                \State Obtain the new cluster assignment $\mathcal{C}'$ using
                Equation~\ref{equation:cluster-update}
                \If{$J(\mathcal{C}') < J(\mathcal{C})$}
                    \State $\mathcal{C} \leftarrow \mathcal{C}'$
                \EndIf
            \EndFor
        \EndFor
    \end{algorithmic}
\end{algorithm}

\subsection{M-step: Estimating the Cluster Centroids}
\sloppy We now look at estimating the set of cluster centroids $\{\ldots \mu_C \ldots \}$ given the cluster assignment $\mathcal{C}$.
Our goal is to minimise the objective function in Equation~\ref{equation:group-objective} while keeping the cluster assignment $\mathcal{C}$ fixed.
Since the $\max$ operator is not differentiable, we use the following differentiable approximation~\cite{P2020Representativity}:
\begin{equation}\label{equation:max-approx}
    \max_{y \in Y} \ g(y) \approx \frac{1}{\phi} \log_e \sum_{y \in Y} \exp{\big(\phi \times g(y)\big)}
\end{equation}
where $\phi$ is a large enough positive constant that amplifies the significance of the largest distance.
Substituting Equation~\ref{equation:max-approx} in Equation~\ref{equation:group-objective} gives us a differentiable approximation for the objective function:
\begin{equation}
    J_\textit{approx} = \sum_{C \in \mathcal{C}} \ \frac{1}{\phi} \log_e \sum_{S \in \mathcal{S}} \exp{\Big(\phi \times \overline{f}(S, C)\Big)}
\end{equation}
We model our optimisation steps along the framework of gradient descent where the intent is to move along the negative gradient. Towards this, we note that the derivative of $J_\textit{approx}$ with respect to $\mu_C$ evaluates to:
\begin{equation}\label{equation:group-gradient}
    \begin{split}
        \diffp{}{{\mu_C}} & J_\textit{approx} = \\
            & -2 \times \frac{\sum\limits_{S \in \mathcal{S}} \Big(w(S, C) \times \frac{1}{|C \cap S|} \times \sum\limits_{x \in C \cap S} (x - \mu_C)\Big)}{\sum\limits_{S \in \mathcal{S}} w(S, C)}
    \end{split}
\end{equation}
where
\begin{equation}\label{equation:w}
    w(S, C) = \exp{\Big(\phi \times \overline{f}(S, C)\Big)}
\end{equation}
Much like in the E-step, regularities in the construction of Equation~\ref{equation:group-gradient} allow for efficient incremental gradient computation.  Equations~\ref{equation:group-gradient} and \ref{equation:w} are together used to update the set of cluster centroids $\{\ldots \mu_C \ldots \}$ within the gradient descent framework:
\begin{equation}\label{equation:gradient-descent}
    \mu_C \leftarrow \mu_C - \eta \diffp{}{{\mu_C}} J_\textit{approx}
\end{equation}
where $\eta \in \mathbb{R}^+$ is the learning rate. The M-step is outlined in Algorithm~\ref{algo:estimate-repr}. To summarise, within each M-step, each cluster centroids, chosen in round-robin fashion, is updated once using the update in Equation~\ref{equation:gradient-descent}.

\begin{algorithm}[ht]
    \caption{Update \label{algo:estimate-repr}}
    \begin{algorithmic}[1]
        \ForAll {$C \in \mathcal{C}$}
            \State Compute $\diffp{}{{\mu_C}} J_\textit{approx}$ using Equations~\ref{equation:group-gradient} and \ref{equation:w}
            \State Perform a single update for $\mu_C$ using Equation~\ref{equation:gradient-descent}
        \EndFor
    \end{algorithmic}
\end{algorithm}

\subsection{Stopping Condition}\label{sec:stopping}
Being initialised using a utilitarian clustering such as one that optimises for the \Kmeans objective, each iteration progressively moves the cluster assignment away from the utilitarian starting point towards cluster-level group representativity fairness. This monotonicity makes the choice of the stopping condition critical to ensure that a good trade-off between the utilitarianism and fairness is achieved within the eventual cluster assignment. In other words, the stopping condition for our iterative procedure is determined by the trade-off between two factors:
\begin{enumerate}[leftmargin=*]
    \item the deterioration in utility, \Ie the classical \Kmeans
    objective, given that our method starts from the cluster assignment generated 
    by the classical \Kmeans algorithm, and
    \item the improvement in our objective function towards group fairness.
\end{enumerate}
It may be noted that both the terms above may be computed intrinsically, \Ie without using any form of external input, and are thus available to the optimisation approach. This allows us to formulate a contrastive stopping condition that can choose good trade-off points in the optimisation. At any iteration, we consider the overall percentage improvement in our fairness objective over the past ten iterations, and the overall percentage deterioration in the \Kmeans objective across the past ten iterations, and break out of the iterative loop when the latter exceeds the former. This is complemented by another stopping condition where we break the iterative process after 200 iterations, if the contrastive stopping condition was not reached earlier.

\section{Empirical Evaluation}

We now detail our empirical evaluation. We start by describing the experimental setup, followed by
results and analyses. All underlying research data and code will be made publicly available upon
acceptance.

\subsection{Experimental Setup}

\subsubsection{Datasets}

The datasets used in our empirical evaluation are based on the publicly available Adult
dataset\footnote{\url{https://archive.ics.uci.edu/ml/datasets/adult}} and CreditCard
dataset\footnote{\url{https://archive.ics.uci.edu/ml/datasets/default+of+credit+card+clients}}
from the UCI repository~\cite{Dua2019UCI}. Both datasets contain data about humans and include
sensitive information such as \textit{race} and \textit{sex}, making them exceedingly popular in the algorithmic
fairness community~\cite{Chhabra2021An, LeQuy2022A}. In case of the Adult dataset, we use
\texttt{workclass}, \texttt{education-num}, \texttt{occupation}, \texttt{capital-gain}, \texttt{capital-loss}, \texttt{hours-per-week} as non-sensitive
attributes. We do not use \texttt{fnlwgt} (refer to Le Quy \Etal~\cite{LeQuy2022A}) and \texttt{education} (which is the same as
\texttt{education-num}). The two sensitive attributes we consider are \texttt{sex} and \texttt{race}. In case of the
CreditCard dataset, we use \texttt{LIMIT\_BAL}, \texttt{PAY\_x} (6 attributes), \texttt{BILL\_AMTx} (6 attributes), \texttt{PAY\_AMTx}
(6 attributes) as the non-sensitive attributes. The sensitive attribute we consider is \texttt{SEX}. The two datasets are processed as follows. Among the non-sensitive attributes,
the continuous ones are standardised (\Ie zero mean and unit variance) on the same lines as
Ghadiri \Etal~\cite{Ghadiri2021Socially}, and the categorical ones are one hot encoded. Table~\ref{tab:datasets} contains information about the processed versions of the datasets used in our
evaluation.

\begin{table}
    \footnotesize
    \centering
    \begin{tabular}{@{}c|cc|c|c@{}}
        \hline
        \multirow{2}{*}{dataset} & \multicolumn{2}{c|}{sensitive attribute} & non-sensitive attributes & \multirow{2}{*}{objects} \\
                                 & name & groups &        & \\
        \hline
        \multirow{2}{*}{Adult}   & \texttt{sex}  &      2 & \multirow{2}{*}{26} & \multirow{2}{*}{30718} \\
                                 & \texttt{race} &      5 &                     &                        \\
        \hline
        CreditCard               & \texttt{SEX}  &      2 &                 77  &                 30000  \\
        \hline
    \end{tabular}
    \caption{Datasets used (after processing)
    \label{tab:datasets}}
\end{table}

\subsubsection{Baselines}

We benchmark our method against the classical \Kmeans algorithm, which also forms our initialisation. Ours being a novel
fairness formulation that has been hitherto unexplored in literature, there are no suitable
state-of-the-art baseline clustering methods to compare against. Abbasi \Etal~\cite{Abbasi2021Fair}
and Ghadiri \Etal~\cite{Ghadiri2021Socially}, being based on group representativity fairness, are related but are optimised for a different fairness objective; we compare our method against
Ghadiri \Etal's Fair-Lloyd~\cite{Ghadiri2021Socially}. Since the available implementation of
Fair-Lloyd\footnote{\url{https://github.com/samirasamadi/SociallyFairKMeans}} can only handle binary sensitive attributes, we do not evaluate it on the Adult dataset with \texttt{race} as the sensitive attribute.
We do not compare with Abbasi \Etal~\cite{Abbasi2021Fair} because it is more suited for the
facility location problem rather than the \Kmeans clustering problem.


\subsubsection{Parameter Configuration}

In all experiments, we set the number of clusters to be generated $k$=5. We run Fair-Lloyd
using its default settings. In case of our method, we set $\phi$=3, and learning rate
$\eta$=0.01 for gradient descent. All numbers reported are averaged over 100 runs with random
initial centroids provided to the classical \Kmeans algorithm (whose generated centroids are
subsequently used as the initial centroids of our method). The numbers reported for Fair-Lloyd
are also averaged over 100 runs. Following Ghadiri \Etal~\cite{Ghadiri2021Socially}'s setup, we
allow Fair-Lloyd and our method to run for 200 iterations for the purpose of our empirical
evaluation.

\subsection{Results}

\subsubsection{Fairness vs Utility}

\begin{table}
    \footnotesize
    \centering
    \begin{tabular}{@{}ccc|rr@{}}
        \hline
        dataset & \multicolumn{1}{m{12mm}}{\centering sensitive attribute} & method & \multicolumn{1}{m{12mm}}{\centering average cluster disparity} & \% change \\
        \hline
        \multirow{6}{*}{Adult}      & \multirow{3}{*}{\texttt{sex}}  & classical \Kmeans & 24.6498 &   0.00\% \\
		                            &                       & Fair-Lloyd        & 25.0157 &  \Red{+1.48\%} \\
		                            &                       & ours              &  9.4066 & \Green{-61.84\%} \\
        \cline{2-5}
                                    & \multirow{3}{*}{\texttt{race}} & classical \Kmeans & 80.8993 &   0.00\% \\
		                            &                       & Fair-Lloyd        &     N/A &      N/A \\
		                            &                       & ours              & 31.9480 & \Green{-60.51\%} \\
		\hline
        \multirow{3}{*}{CreditCard} & \multirow{3}{*}{\texttt{SEX}}  & classical \Kmeans & 21.6800 &   0.00\% \\
		                            &                       & Fair-Lloyd        & 16.8425 & \Green{-22.31\%} \\
		                            &                       & ours              &  4.8128 & \Green{-77.80\%} \\
        \hline
    \end{tabular}
    \caption{Average cluster disparity (lower is better). The \% change column shows the increase (in \Red{red}) or decrease (in \Green{green}) in the average cluster disparity with the classical \Kmeans algorithm as the reference.
    \label{tab:acd}}
\end{table}

\begin{table}
    \footnotesize
    \centering
    \begin{tabular}{@{}ccc|rr@{}}
        \hline
        dataset & \multicolumn{1}{m{12mm}}{\centering sensitive attribute} & method & \multicolumn{1}{m{12mm}}{\centering \Kmeans objective} & \% change \\
        \hline
        \multirow{6}{*}{Adult}      & \multirow{3}{*}{\texttt{sex}}  & classical \Kmeans & 2.0849 &   0.00\% \\
                                    &                       & Fair-Lloyd        & 2.3548 & +12.95\% \\
                                    &                       & ours              & 2.5833 & +23.91\% \\
        \cline{2-5}
                                    & \multirow{3}{*}{\texttt{race}} & classical \Kmeans & 2.0860 &   0.00\% \\
		                            &                       & Fair-Lloyd        &     N/A &      N/A \\
                                    &                       & ours              & 2.3299 & +11.69\% \\
        \hline
        \multirow{3}{*}{CreditCard} & \multirow{3}{*}{\texttt{SEX}}  & classical \Kmeans & 8.6176 &   0.00\% \\
                                    &                       & Fair-Lloyd        & 8.5498 &  -0.79\% \\
                                    &                       & ours              & 9.8836 & +14.69\% \\
        \hline
    \end{tabular}
    \caption{\Kmeans objective (lower is better). The \% change column shows the increase/decrease in the \Kmeans objective with the classical \Kmeans algorithm as the reference.
    \label{tab:kmeans-objective}}
\end{table}

Our method optimises for a fairer objective function different from the classical utilitarian one.
It is widely accepted that increase in fairness almost always causes decrease in utility;
it would be of interest to look at the fairness gains obtained due to our method and the
corresponding loss in utility. Here, we quantify unfairness with our evaluation metric -- average
cluster disparity (Equation~\ref{equation:acd}), and utility with the \Kmeans objective\footnote{We normalise the original \Kmeans objective here, \Ie divide Equation~\ref{equation:kmeans} by the number of objects in the dataset.}. From Tables~\ref{tab:acd} and \ref{tab:kmeans-objective}, we see that our method is able to significantly reduce
the unfairness (reduction of 61.84\%, 60.51\%, 77.80\% over the classical \Kmeans algorithm on our
datasets; 66.72\% on average) at the cost of a relatively smaller increase in the utilitarian
objective (increase of 23.91\%, 11.69\%, 14.69\% over the classical \Kmeans algorithm; 16.76\% on
average). Fair-Lloyd, on the other hand, does not perform as well on this fairness metric (average
reduction of 10\%). Evidently, our method outperforms the baselines on our cluster-level group
fairness evaluation metric, \Ie average cluster disparity.

\subsubsection{Fairness Objectives}

\begin{table}
    \footnotesize
    \centering
    \begin{tabular}{@{}ccc|rr@{}}
        \hline
        dataset & \multicolumn{1}{m{12mm}}{\centering sensitive attribute} & method & \multicolumn{1}{m{12mm}}{\centering our objective} & \% change \\
        \hline
        \multirow{6}{*}{Adult}      & \multirow{3}{*}{\texttt{sex}}  & classical \Kmeans &  3.1302 &   0.00\% \\
		                            &                       & Fair-Lloyd        &  3.5403 & \Red{+13.10\%} \\
		                            &                       & ours              &  2.1299 & \Green{-31.96\%} \\
		\cline{2-5}
		                            & \multirow{3}{*}{\texttt{race}} & classical \Kmeans &  4.1148 &   0.00\% \\
		                            &                       & Fair-Lloyd        &     N/A &      N/A \\
		                            &                       & ours              &  2.7868 & \Green{-32.27\%} \\
        \hline
        \multirow{3}{*}{CreditCard} & \multirow{3}{*}{\texttt{SEX}}  & classical \Kmeans & 43.8224 &   0.00\% \\
		                            &                       & Fair-Lloyd        & 39.6524 &  \Green{-9.52\%} \\
		                            &                       & ours              & 14.3313 & \Green{-67.30\%} \\
        \hline
    \end{tabular}
    \caption{Our objective (lower is better). The \% change column shows the increase (in \Red{red}) or decrease (in \Green{green}) in our objective with the classical \Kmeans algorithm as the reference.}
    \label{tab:our-objective}
\end{table}

\begin{table}
    \footnotesize
    \centering
    \begin{tabular}{@{}ccc|rr@{}}
        \hline
        dataset & \multicolumn{1}{m{12mm}}{\centering sensitive attribute} & method & \multicolumn{1}{m{12mm}}{\centering fair \Kmeans objective} & \% change \\
        \hline
        \multirow{6}{*}{Adult}      & \multirow{3}{*}{\texttt{sex}}  & classical \Kmeans &  2.2473 &   0.00\% \\
		                            &                       & Fair-Lloyd        &  2.5453 & +13.26\% \\
		                            &                       & ours              &  2.7811 & +23.75\% \\
        \cline{2-5}
                                    & \multirow{3}{*}{\texttt{race}} & classical \Kmeans &  2.6185 &   0.00\% \\
		                            &                       & Fair-Lloyd        &     N/A &      N/A \\
		                            &                       & ours              &  2.8538 &  +8.99\% \\
        \hline
        \multirow{3}{*}{CreditCard} & \multirow{3}{*}{\texttt{SEX}}  & classical \Kmeans &  8.6489 &   0.00\% \\
		                            &                       & Fair-Lloyd        &  8.5508 &  -1.13\% \\
		                            &                       & ours              & 10.1417 & +17.26\% \\
        \hline
    \end{tabular}
    \caption{Fair \Kmeans objective~\cite{Abbasi2021Fair, Ghadiri2021Socially} (lower is better). The \% change column shows the increase/decrease in the fair \Kmeans objective with the classical \Kmeans algorithm as the reference.
    \label{tab:fair-kmeans-objective}}
\end{table}

In the interest of comparing across the two threads of representativity fairness, we look at how our method
compares with the baselines on the two fairness objectives:
\begin{enumerate*}[label=(\roman*)]
    \item our objective\footnote{We normalise our objective here, \Ie divide Equation~\ref{equation:group-objective} by the number of clusters $k$.} (Equation~\ref{equation:group-objective}), and
    \item fair \Kmeans objective~\cite{Abbasi2021Fair, Ghadiri2021Socially} (Equation~\ref{equation:fair-kmeans})
\end{enumerate*}.
In Table~\ref{tab:our-objective}, we see that our method does indeed improve the representativity of
the worst-off sensitive groups (reduction of 31.96\%, 32.27\%, 67.30\% over the classical \Kmeans algorithm; 43.84\% on average) which indicates that our method is moving in
the right direction. The Fair-Lloyd baseline does not perform as well; this is acceptable as it is not designed for cluster-level group fairness. On the other hand, our method causes the worst-off
sensitive group at the cluster assignment level to have a worse representativity than the classical \Kmeans algorithm, as seen in Table~\ref{tab:fair-kmeans-objective}. While at first glance this may seem to be a deficiency of our method, we note that any improvement in the representativity of this worst-off group would potentially result in another sensitive group that is the worst-off in some cluster being further disadvantaged in order to improve the representativity of the former group. This would be unacceptable in cases where more than one sensitive groups are historically disadvantaged (\Eg black and American Indians in case of \textit{race}) and benefitting one would result in disadvantaging the other.

\subsubsection{Disparity Trends}

\begin{figure}
    \centering
    \includegraphics{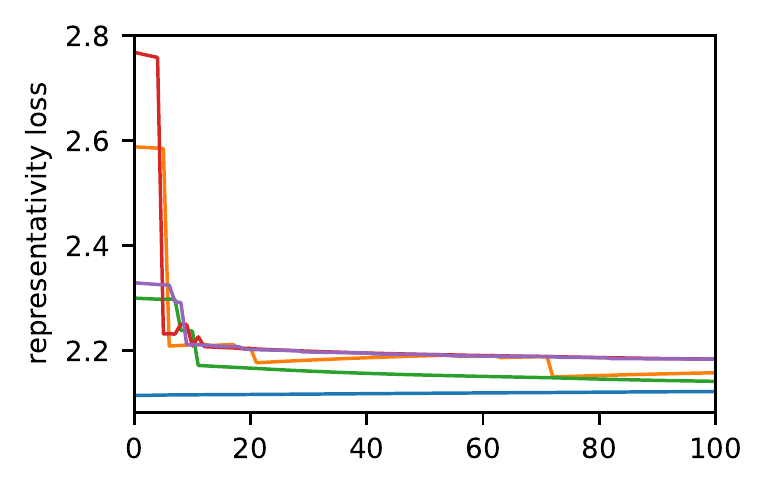}
    \caption{Trends in representativity loss $\overline{f}(.,.)$ of the 5 sensitive groups for the Adult dataset with \texttt{race} as the sensitive attribute (\texttt{random\_state}=3, cluster 4) over 100 iterations. Lower is better.
    \label{fig:trends}}
\end{figure}

In order to further observe the behaviour of our method, we look at how the representativity
losses of the different sensitive groups vary within a cluster over iterations. As an
illustration, Figure~\ref{fig:trends} shows the trends in cluster 4 that was initialised with
\texttt{random\_state}=3 on the Adult dataset with \texttt{race} as the sensitive attribute.
We notice that over iterations
\begin{enumerate*}[label=(\roman*)]
    \item the disparity in the representativities of the best-off and worst-off sensitive groups decreases, and
    \item the representativity loss of the worst-off sensitive group decreases
\end{enumerate*},
which is what we expect. Also note in this figure that the worst-off sensitive group in the cluster changes
over iterations; this is a trend that we observe to generally hold for clusters.

\section{Conclusion and Future Work}

In this paper, we introduced a new formulation of fairness for clustering with respect to representativity of sensitive groups at the cluster-level. Against the backdrop of much recent enthusiasm in using representativity-oriented notions of fairness in clustering, we outlined issues engendered by simplistic aggregations of representativities across clusters, and how our notion alleviates such problems. We proposed a new post-processing based clustering formulation that is able to improve upon existing cluster assignments by iteratively modifying them towards cluster-level group representativity fairness. Given the novel form of our fairness notion, we introduced a new metric that captures disparity between the representativity of groups at the cluster level.
Our experiments on the Adult and CreditCard datasets demonstrate the effectiveness of our method towards achieving high levels of cluster-level group representativity fairness at low impact to the popular utilitarian cluster coherence metric used within \Kmeans. 

\subsection{Future Work}

One future direction for this work would be to extend our method to incorporate multiple sensitive attributes together, and also to consider numeric sensitive attributes such as \textit{age}, an attribute on which discrimination is well-understood within healthcare scenarios. Another direction would be to extend this notion to other clustering paradigms such as hierarchical clustering and spectral clustering where data models depart significantly from the relational model we assumed in our present work.

\section*{Acknowledgement}

This project has received funding from the European Union’s Horizon 2020 research and innovation programme under the Marie Skłodowska-Curie grant agreement No 945231; and the Department for the Economy in Northern Ireland. We are grateful for use of the computing resources from the Northern Ireland High Performance Computing (NI-HPC) service funded by EPSRC (EP/T022175).

\printbibliography

\end{document}